\documentclass[runningheads]{llncs}
\usepackage{graphicx}
\usepackage{paralist}

\usepackage{tikz}
\usepackage{comment}
\usepackage{amsmath,amssymb} 
\usepackage{color}

\usepackage{amsfonts}       
\usepackage[cal=boondox]{mathalfa}

\usepackage[accsupp]{axessibility}  

\usepackage[width=122mm,left=12mm,paperwidth=146mm,height=193mm,top=12mm,paperheight=217mm]{geometry}

\begin{document}
\pagestyle{headings}
\mainmatter
\def\ECCVSubNumber{7438}  

\title{Detection and Mapping of Specular Surfaces Using Multibounce Lidar Returns} 

\titlerunning{Detection and Mapping of Specular Surfaces}
%
\author{Connor Henley \inst{1,2} \and
Siddharth Somasundaram\inst{1} \and
Joseph Hollmann \inst{2} \and
Ramesh Raskar \inst{1}}
\authorrunning{C. Henley et al.}
%

\institute{Massachusetts Institute of Technology\\
\email{\{co24401,sidsoma,raskar\}@mit.edu}\\
\and
The Charles Stark Draper Laboratory\\
\email{jhollmann@draper.com}}
\maketitle

\begin{abstract}
We propose methods that use specular, multibounce lidar returns to detect and map specular surfaces that might be invisible to conventional lidar systems that rely on direct, single-scatter returns.  We derive expressions that relate the time- and angle-of-arrival of these multibounce returns to scattering points on the specular surface, and then use these expressions to formulate techniques for retrieving specular surface geometry when the scene is scanned by a single beam or illuminated with a multi-beam flash.  We also consider the special case of transparent specular surfaces, for which surface reflections can be mixed together with light that scatters off of objects lying behind the surface.
\end{abstract}

\section{Introduction}




Although lidar is widely used for mapping the 3D geometry of surfaces, the technology has historically been challenged by \emph{specular}, or mirror-like, surfaces that typically scatter very little light directly back to the receiver.  This inability to detect and localize specular surfaces can result in the failure to detect navigational obstacles like mirrors and windows, or hazards such as wet or icy patches on the ground.  It can also result in incomplete scans of cityscapes or man-made interior environments in which glass and metal surfaces are relatively common, and in the complete inability to digitize artifacts that are made of glass or that present a polished metal or chrome finish.

\begin{figure}[!t]
\centering
    \includegraphics[width=\textwidth]{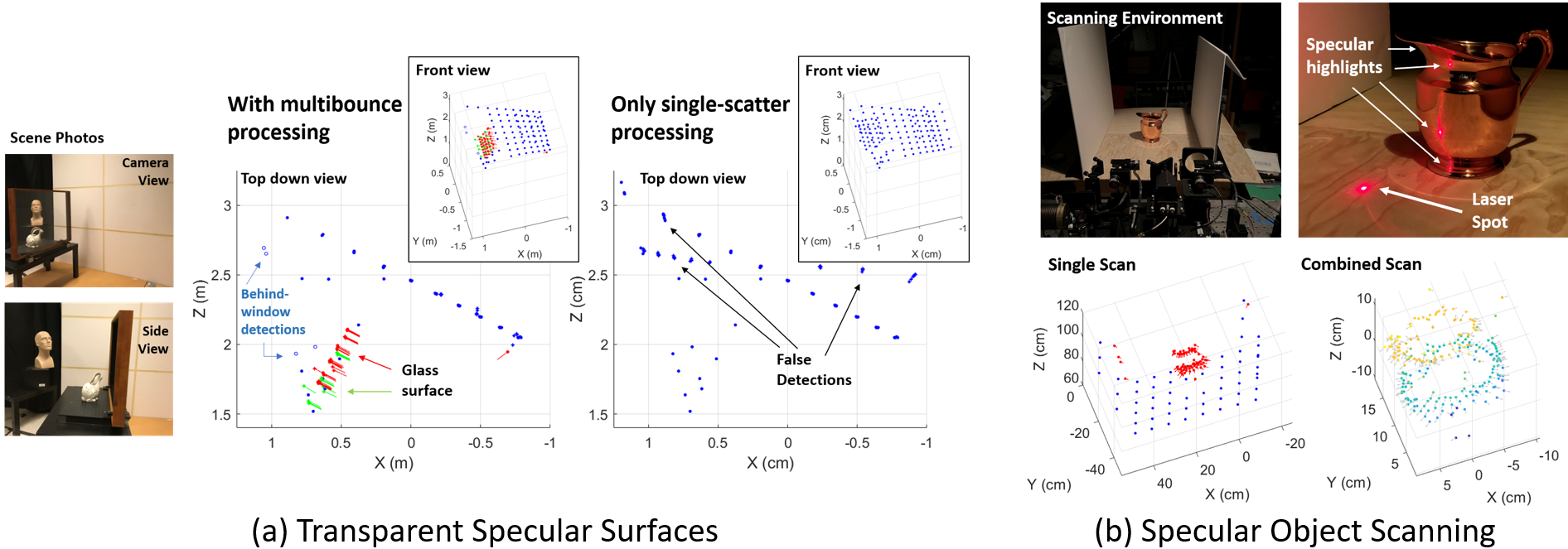}
    \caption{(a) We acquire the shape of a glass window (green and red points) and the scene surrounding it (blue points), which includes objects behind the window (white points, photos on left). When single-scatter paths are assumed (right), the window surface cannot be retrieved and numerous false points are generated. (b) We scan the shape of a copper pitcher (top) by directly illuminating a sequence of 60 laser spots on nearby surfaces and measuring the time-of-flight of two-bounce returns that reflect off of the pitcher (results bottom left).  We repeat the scan four times, rotating the pitcher $90^\circ$ between scans, to acquire the full shape (bottom right).}
    \label{fig:teaser}
\end{figure}

The range to a surface is typically computed using the lidar range equation, which is only valid for single-scatter time-of-flight measurements.  Thus, when the direct, single-scatter return from a specular surface is too faint to detect, it becomes impossible to compute the range to that surface via conventional means.  Nevertheless, the presence of specular surfaces is often revealed by intense \emph{multibounce} returns.  For instance, when one directly illuminates a \emph{diffusely} reflecting surface, nearby specular surfaces may produce mirror images of the true laser spot (also referred to as \emph{highlights}) that appear just as bright as the original.  Alternatively, when a specular surface is illuminated directly, the beam is often deflected such that it lands on a diffusely reflecting surface nearby.

In this work we demonstrate that multibounce returns are both an important cue that reveals the presence of specular surfaces, as well as an information source that can be used to estimate a specular reflector's shape.  We review the geometry of multibounce returns that are encountered in scenes that contain both diffuse and specular reflectors.  Using our knowledge of this geometry, we propose criteria for unambiguously detecting the presence of specular multibounce signals, as well as a set of equations that relates the time- and angle-of-arrival of multibounce impulses to the position and orientation of points on the specular surface.  

We apply our findings in a series of experiments in which we use multibounce lidar measurements to acquire the 3D shape of various specular objects.  These objects include planar reflectors such as mirrors and windows, as well as a polished metal object with a freeform shape.  For the special case of transparent surfaces, we propose criteria that can be used to distinguish between multibounce reflections off of the surface, and single-scatter returns from objects that lie behind the surface.

Our methods can be implemented using any lidar system for which the transmit and receive axes can be steered independently, or that uses a wide FOV receiver such as a single-photon avalanche diode (SPAD) array.  For most of our experiments we use a pencil-beam illumination source that must be scanned to acquire a point cloud of the full scene.  However we also propose an algorithm that can be used to map the surface of planar reflectors when the scene is illuminated by many beams simultaneously.  Multi-beam illumination enables faster scene acquisition, and is employed by several commercial lidar scanners \cite{ipadpro}.

\section{Related Work}

\subsubsection{Detecting Mirrors and Windows Using Lidar}

Specular surfaces typically reflect most light away from the lidar receiver, resulting in very faint direct returns.  When specular reflections \emph{are} directed towards the receiver, the signal can be so intense that it saturates one's detectors (this event is sometimes called ``glare").  The presence of mirrors in a scene can also produce false detections of the \emph{mirror images} of real objects, which typically appear behind the mirror.

To overcome these challenges, Diosi and Kleeman \cite{diosi2004} use an ultrasonic scanner to detect specular surfaces that lidar scanners can't see.  Yang et al. \cite{yang2010} infer that framed gaps in detected surfaces likely contain mirrors or windows.  After classifying mirrors in this way, they identify mirror image detections and reflect them across the mirror plane to the position of the true object.  Foster et al. use sparse glare events to detect specular surfaces, but accumulate glare information over time in an occupancy grid as their laser scanner moves through a space \cite{foster2013}.  Like Yang et al., Tibebu et al. \cite{tibebu2021} search for frames that are indicative of windows.  However, to avoid detecting windowless holes, they use measurable pulse broadening caused by transmission through glass as a second heuristic to make their window detector more robust.


Unlike this previous work, our method does not rely on contextual cues like frame or mirror image detection that might be produced by non-specular scenes, we do not rely on rare glare events, and we do not require additional sensing modalities like ultrasound.  The work that comes closest to ours was a speculative report by Raskar and Davis \cite{raskar2008}, who suggested that the time-of-flight of two-bounce returns could be used to localize specular surfaces, but did not demonstrate their method, and did not (as we do) consider the scenario where the specular surface is illuminated directly.  Henley et al. \cite{henley2021} and Tsai et al. \cite{tsai2016} estimate scene shape from two-bounce time-of-flight measurements, but neither consider the special conditions imposed by specular reflectors.

\subsubsection{Specular Surface Estimation Using Cameras}

There is an extensive body of literature that investigates methods for estimating the geometry of specular surfaces using conventional cameras that cannot capture time-of-flight information.  A relatively recent review of this research was provided by Ihrke et al. \cite{ihrke2010}.  A camera that observes a specular surface will see a distorted (if the mirror is curved) reflection of the scene that surrounds the surface.  In most camera-based methods for specular geometry estimation, features in the distorted, reflected image are matched to features in the true scene.  If the positions of the camera and true feature are known, then the \emph{depth} and \emph{surface normal} of the surface point that reflects the distorted feature can be determined up to a one-dimensional ambiguity.  This ambiguity can be resolved in a variety of ways.  In \emph{shape from specularity} methods, the reflected scene features are point sources that produce specular highlights in the camera image \cite{zisserman1989}\cite{blake1988}.  In \emph{shape from distortion} methods, a camera observes how specular reflection distorts the reflected image of a reference pattern \cite{savarese2005}\cite{whelan2018}\cite{bonfort2003}.  In \emph{specular flow} methods a camera observes how the distorted reflection of an \emph{uncalibrated} scene appears to move across the surface of a specular object as the camera, object, or scene are moved \cite{roth2006}.





Because we observe the specular reflection of laser spots, our method can be interpreted as a \emph{shape from specularity} technique that uses time-of-flight constraints to resolve the depth-normal ambiguity.  In future work, it would be interesting to explore the combination of time-of-flight constraints with other imaging strategies such as \emph{shape from distortion} or \emph{specular flow}.  

\section{Methods}

\subsection{Geometry of Specular Multibounce Signals}
\label{sec:geom}

\begin{figure*}[!t]
\centering
    \includegraphics[width=0.8\textwidth]{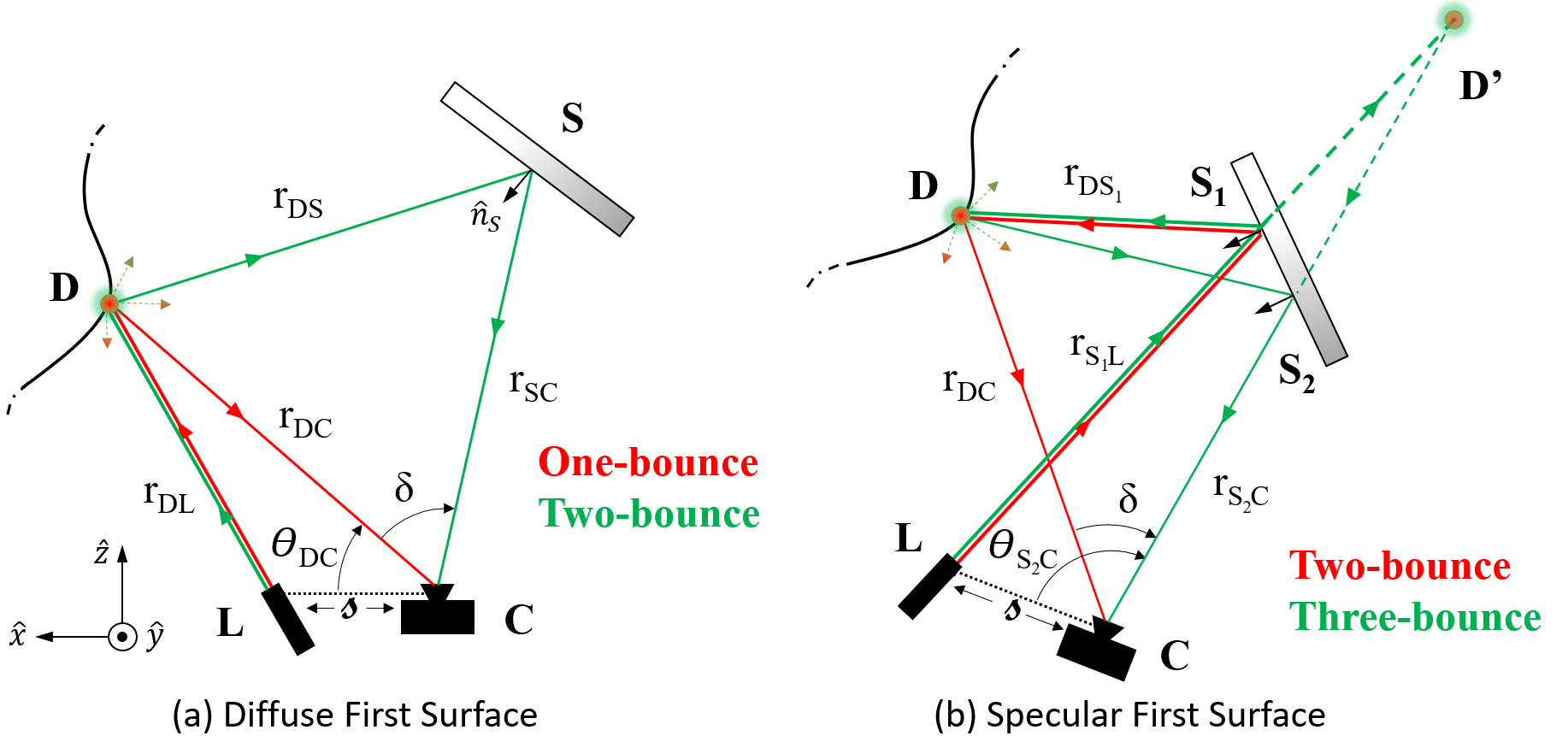}
    \caption{(a) A diffusely reflecting surface is illuminated directly at $D$.  The camera observes one-bounce returns incident from $D$ and two-bounce returns incident from point $S$ on the mirror surface.  (b) The specular surface is illuminated directly (at $S_1$) and the single-bounce return is too faint to observe.  The camera observes two-bounce returns incident from $D$ and three-bounce returns incident from $S_2$.}
    \label{fig:mirror_geom}
\end{figure*}

Consider the scenario illustrated in Fig. \ref{fig:mirror_geom}, where a mirror has been placed in an environment that is otherwise composed of diffusely reflecting surfaces.  The mirror is a perfect specular reflector, which means that it will reflect all light from an incident beam in a single direction that is determined by the light's direction of incidence and the mirror's surface normal orientation.  Our lidar system consists of a transmitter at position $L$ that emits a focused, pulsed beam in a single direction.  A receiver array is placed at position $C$, which may be displaced from L by a baseline distance $\mathcal{s}$. 

In this scenario the transmitter can illuminate either a diffuse reflector or the mirror directly.  These two cases introduce distinct multibounce geometries that must be treated separately.  In either case, our goal is to compute the position of all directly and indirectly illuminated points in the scene using the time-of-flight and angle-of-arrival of all observable direct and multibounce returns.

\subsubsection{First surface is a diffuse reflector.}  In the first case, shown in Fig. \ref{fig:mirror_geom}(a), the transmitter directly illuminates a point $D$ on the diffusely reflecting surface.  The receiver subsequently observes two laser spots: the \emph{true} spot, which (correctly) appears to be located at $D$, and a \emph{mirror image} of that spot.  Light from the true spot has propagated along the single-bounce path $LDC$, whereas light that appears to originate from the spot's mirror image has propagated along the two-bounce path $LDSC$.  Here $S$ is the point at which light reflects off of the mirror.  We use a method that was originally introduced in \cite{henley2021} to estimate the positions of $D$ and $S$ from the times- and angles-of-arrival of the one- and two-bounce returns.  Angular positions are described by a clockwise rotation by $\theta$ about $\hat{y}$ followed by a clockwise rotation by $\phi$ about $\hat{x}$.


Light incident from the direction of the true spot $D$ will arrive at the angle ($\theta_{DC}$, $\phi_{DC}$) at time $t_1 = \frac{1}{c}\left(r_{DL}+r_{DC}\right)$, where $r_{DL}$ and $r_{DC}$ are the distances to $D$ from $L$ and $C$, respectively.  The distance to $r_{DC}$ can be computed using the following bi-static range equation:

\begin{equation}
    \label{eq:rdc}
    r_{DC} = \frac{1}{2} \frac{c^2 t_1^2 - \mathcal{s}^2}{c t_1 - \mathcal{s}cos(\theta_{DC})}.
\end{equation}

Light incident from the direction of $S$ will arrive at the angle ($\theta_{SC}$, $\phi_{SC}$) at time $t_2 = \frac{1}{c}\left(r_{DL}+r_{DS}+r_{SC}\right)$.  Here $r_{DS}$ is the distance from $S$ to $D$ and $r_{SC}$ is the distance from $C$ to $S$.  From simple arithmetic we see that $r_{DS} = c(t_2 - t_1) + r_{DC} - r_{SC}$.  We plug this expression into the law of cosines for the triangle DCS to obtain

\begin{equation}
    \label{eq:rsc}
    r_{SC} = \frac{c}{2} \cdot \frac{\Delta t_{12} [ \Delta t_{12} + \frac{2 r_{DC}}{c} ]}{\Delta t_{12} + (1 - \cos{\delta})\frac{r_{DC}}{c}}.
\end{equation}

Here $\Delta t_{12} = t_2 - t_1$ and $\delta$ is the apparent angular separation of the true spot and its mirror image, with $\cos{\delta} = \cos{\theta_{DC}}\cos{\theta_{SC}} + \sin{\theta_{DC}}\sin{\theta_{SC}}\cos{\left(\phi_{DC} - \phi_{SC}\right)}$.

We have now completely determined the positions of $S$ and $D$.  We can additionally compute the surface normal at $S$ which, by the law of reflection, must be the \emph{normalized bisector} of the angle formed by line segments $DS$ and $CS$.

\subsubsection{First surface is a specular reflector.}  In the second case, shown in Fig. \ref{fig:mirror_geom}(b), the mirror is illuminated directly at $S_1$.  Because the mirror is a perfect specular reflector, no light that scatters at $S_1$ will travel directly back to the receiver, and so we do not observe a one-bounce return.  Instead, all light in the beam is deflected such that it illuminates a spot $D$ on a diffusely reflecting surface.  This laser spot is visible to the receiver.  Once again, however, the receiver \emph{also} sees a mirror image of the true laser spot.  The mirror image of $D$ appears to lie behind\footnote{Assuming that the mirror is not strongly concave} the mirror at point $D'$.  This time, light that arrives from the true spot has traveled along the two-bounce path $LS_1DC$, and light that arrives from the spot's mirror image follows the \emph{three}-bounce path $LS_1DS_2C$.

Here we show that it is possible to compute the positions of $D$, $S_1$ and $S_2$ \emph{if} $S_1$ and $S_2$ lie on the same plane\footnote{More precisely, if there is a single plane that is tangent to the mirror surface at both $S_1$ and $S_2$.}.  This condition is automatically satisfied if the specular surface is itself a plane, or if the baseline $\mathcal{s}=0$ (in which case $S_1 = S_2$).  

Our derivation hinges upon the observation that light that has actually followed the three-bounce trajectory $LS_1DS_2C$ will \emph{appear}, from the perspective of the receiver, to have followed the \emph{one}-bounce trajectory $LD'C$, which includes a single scattering event at $D'$.  We can thus compute the apparent range to $D'$ from the three-bounce travel time $t_3 = \frac{1}{c}\left(r_{S_1L} + r_{DS_1} + r_{DS_2} + r_{S_2C}\right) = \frac{1}{c}\left(r_{D'L} + r_{D'C}\right)$.  Using the range equation from Eq. \ref{eq:rdc}, we obtain

\begin{equation}
    \label{eq:rdpc}
    r_{D'C} = r_{DS_2} + r_{S_2C} = \frac{1}{2} \frac{c^2 t_3^2 - \mathcal{s}^2}{c t_3 - \mathcal{s}cos(\theta_{S_2C})}.
\end{equation}

The range to the illusory point $D'$ is useful because it can be directly related to the range of the \emph{true} point $D$ using the following expression:

\begin{equation}
    \label{eq:rdc3b}
    r_{DC} = r_{D'C} - c(t_3 - t_2),
\end{equation}

where $t_2 = \frac{1}{c}\left(r_{S_1L} + r_{DS_1} + r_{DC}\right)$ is the two-bounce travel time.  Once we've obtained $r_{DC}$ we compute the range to $S_2$ by substituting $r_{DS_2} = c(t_3 - t_2) + r_{DC} - r_{S_2C}$ into the law of cosines for the triangle $DCS_2$ to obtain

\begin{equation}
    \label{eq:rs2c}
    r_{S_2C} = \frac{c}{2} \cdot \frac{\Delta t_{23} [ \Delta t_{23} + \frac{2 r_{DC}}{c} ]}{\Delta t_{23} + (1 - \cos{\delta})\frac{r_{DC}}{c}},
\end{equation}

where $\Delta t_{23} = t_3 - t_2$ and $\delta$ once again refers to the apparent angular separation of the true spot and its mirror image.  We note that if there are multiple specular surfaces in the scene then there may be multiple mirror image spots that are visible to the receiver.  Only light from one of those mirror images can be used to compute $r_{DC}$ via Eq. \ref{eq:rdc3b}.  This will be the mirror image that appears to lie along the transmitted beam  (in the direction of $S_1L$).  However, the range to the reflection points on other specular surfaces can be computed once $r_{DC}$ is known.  This is accomplished by plugging the angle-of-arrival and three-bounce time-of-flight associated with these other mirror image spots into Eq. \ref{eq:rs2c}.

If the direction of the transmitted beam is known then we can also compute the position of the directly illuminated point $S_1$.  We begin by computing the apparent distance $r_{D'L}$ from the laser to $D'$:

\begin{equation}
    \label{eq:rdpl}
    r_{D'L} = ct_3 - r_{D'C} = ct_2 - r_{DC}. 
\end{equation}

The distance from $L$ to $S_1$ can then be computed using the identity $r_{DS_1} = r_{D'L} - r_{S_1L}$ and the law of cosines from the triangle $S_1LD$.  This distance is

\begin{equation}
    \label{eq:rls1}
    r_{LS_1} = \frac{1}{2}\frac{r_{D'L}^2 - r_{DL}^2}{r_{D'L} - r_{DL}\cos{\delta_L}},
\end{equation}

where $r_{DL} = ||D-L||_2$ and $\cos{\delta_L} = \frac{1}{r_{DL}}\left(D - L\right) \cdot \hat{d}_{S_1L}$.  Here $\hat{d}_{S_1L}$ is the direction of the transmitted beam.  Finally, from the law of reflection, we can compute the surface normal vectors at $S_1$ and $S_2$ once $D$, $S_1$ and $S_2$ are known.

\subsubsection{Identifying the true spot and the first surface.}  Although we have derived expressions that could in theory be used to compute the positions of all diffuse and specular scattering points within the scene, there are two ambiguities that need to be resolved before these expressions can be applied.  First, we must determine whether the directly illuminated surface is a diffuse or a specular reflector.  Doing so from raw measurements is not as straightforward as it might seem because in either case our receiver will see at least two spots, and one of these spots will appear to lie along the transmitter's boresight.  Second, regardless of which surface was illuminated first, we must also determine which observed spot is the true laser spot and which are mirror images.

Resolving the second ambiguity is always straightforward if we have time-of-flight information.  \textbf{Light from the true spot always arrives \emph{before} the light from the mirror images.}  This can be confirmed by inspecting Fig. \ref{fig:mirror_geom}.  In this figure all propagation paths travel through $D$.  However, light from the true spot travels directly from $D$ to $C$, whereas light from the mirror images must follow a longer, indirect path that includes an additional reflection.  The first-surface ambiguity can be resolved once the true spot has been identified.  \textbf{If the true spot lies along the transmitter boresight, then a diffuse reflector was illuminated first.  If it doesn't, then the specular surface was illuminated first.} Once both ambiguities have been resolved, the appropriate range equations can be applied to determine the scattering points.

It is worth noting that we are only able to disambiguate these two cases because we have time-of-flight information.  If we had instead been viewing the scene with a regular camera that only measured angles and intensities, then there would be no direct way to determine which spot was the true spot and which surface was illuminated first.  Instead, we would have to rely on contextual cues to determine, for instance, which pixels seemed to have mirrors in them.

\subsection{Transparent Specular Surfaces}
\label{sec:window}


\begin{figure}[!t]
\centering
    \includegraphics[width=\textwidth]{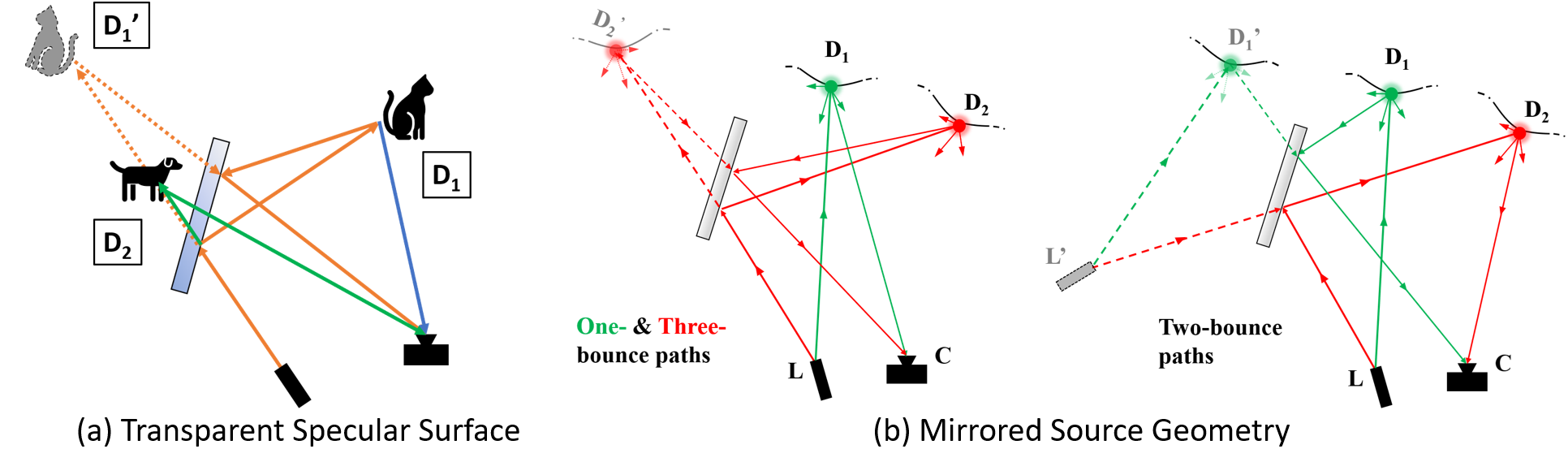}
    \caption{(a) When a transparent specular surface is illuminated directly, multiple laser spots may be detected along the beam vector---a true spot $D_2$ that lies behind the surface, and a mirror image ($D_1'$) of a spot $D_1$ that lies in front of the surface. (b) Two-bounce paths (right) appear to originate at $L'$, a mirror image of the true source.  One- and three-bounce paths (left) appear to originate at the true source $L$.}
    \label{fig:winMB}
\end{figure}


Many of the most common specular surfaces---glass windows, for example---are \emph{partially transparent}.  Transparency introduces an additional source of ambiguity into our measurements, which is highlighted in Fig. \ref{fig:winMB}(a).  Here we see that, as was true for the mirror surface, three-bounce returns will typically resemble spots that lie behind the specular surface.  When that surface is also transparent, however, these three-bounce returns may be mixed together with one-bounce returns that have scattered diffusely off of something that lies behind (or on) the specular surface.  In this section we propose several criteria that can be used to detect and disambiguate these two kinds of signals.


\paragraph{Multiple Detections Along Beam.}  Our disambiguation logic is triggered when we detect multiple spots lying along the transmitted beam vector.  When this occurs, it suggests that we may have detected the mirror image of a true spot that lies in the scene in front of the window, in addition to a one-bounce return from a true spot that lies behind the window or on its surface.


\paragraph{Comparison to Two-bounce Travel Time.}  When the window is illuminated directly, light that arrives from the true spot in front of the window follows a \emph{two-bounce} path, and the spot will not lie on the beam vector.  As was explained in Sec. \ref{sec:geom}, the mirror image in this scenario must correspond to a three-bounce return, and the associated three-bounce time-of-flight must be \emph{greater} than the two-bounce time-of-flight from the true spot.  Thus, if the time-of-flight associated with any of the spots that lie along the beam vector is \emph{less} than the two-bounce time-of-flight, we deduce that they cannot be mirror image returns and, thus, must be single-scatter returns from on or behind the window surface.


\paragraph{Multiple Reflections Diminishes Intensity.}  If the time-of-flight associated with more than one in-beam spot is greater than the two-bounce time-of-flight, we infer that the \emph{dimmest} spot corresponds to the mirror image.  The reasoning behind this is that three-bounce signals are reflected twice by the window, whereas one-bounce signals are transmitted twice.  Common transparent, specular materials typically transmit more light than they reflect, although this assumption can break down at glancing incidence angles.  To choose the dimmest spot, we rank in-beam detections by their range-adjusted intensity ($\sim r^2I$).  For this computation, we use the \emph{apparent} range of each spot, computed using Eq. \ref{eq:rdc}.


\subsection{Illumination with Multiple Beams}
\label{sec:MB}




It is straightforward to employ the concepts introduced in Secs. \ref{sec:geom} and \ref{sec:window} to acquire specular-surface geometry \emph{if} the scene can be scanned by transmitting a single beam at a time. Unfortunately, the time required to do so may be unacceptably long for some applications.  In this section we propose an algorithm that permits the mapping of specular surface geometry without any knowledge of spot-to-beam associations, and thus can be implemented when the scene is illuminated by a multi-beam flash.


\subsubsection{Mirrored Source Geometry}

If a transmitter emits a pulse at time $t$ in a scene that contains a flat mirror, then a receiver that is pointed at the mirror will observe a mirror image of the transmitter that \emph{also appears} to emit a pulse at $t$.  The direction that the mirrored pulse propagates will be flipped across the mirror plane.  This geometry is visualized in Fig. \ref{fig:winMB}(b), where the true source is at $L$, and its mirror image is at $L'$.  The position of $L'$ is significant because the plane of the mirror perfectly bisects the line segment $LL'$, and the plane's surface normal is parallel to $LL'$.  This means that estimating the position of $L'$ (relative to $L$) is equivalent to estimating the mirror plane.  This is the key principle that underlies our multi-beam surface estimation algorithm.


The one-, two-, and three-bounce signals described in previous sections can also be interpreted under a mirrored space model.  From the perspective of the receiver, one- and three-bounce returns appear to originate from the true source $L$. \emph{Two}-bounce returns appear to originate from the \emph{mirrored} source, at $L'$.  From Fig. \ref{fig:winMB}(b), we see that the order of the bounces determines where the scattering event will appear to occur.  If light bounces off of the mirror first, then the scattering event appears to occur in the true space at $D$.  If it bounces off a diffuse reflector first, then light appears to scatter in the mirrored space at $D'$.

\subsubsection{Source Localization Using Multilateration}

Because spots produced by one- or three-bounce light paths appear to originate from the true source, they must also lie along the axis of one of the true, transmitted beams.  Spots that result from two-bounce light paths, on the other hand, will not in general lie along a transmitted beam vector.  Because of this it is straightforward to determine which detected spots correspond to two-bounce light paths.  An image of spot detections classified in a multi-beam data collection is shown in Fig. \ref{fig:mb_res}(b).


If we could compute the apparent positions of at least three two-bounce spots, then we could compute the ranges $r_{DL'} = ct_2 - r_{DC}$ or $r_{D'L'} = ct_2 - r_{D'C}$, and determine $L'$ by solving a multilateration problem.  Although we cannot compute these positions directly without knowledge of spot-to-beam associations, we can approximate them.  From one-bounce returns, we compute the positions of several points in the true space using Eq. \ref{eq:rdc}.  We also compute the positions of a number of points in the mirrored space from three-bounce returns, using Eq. \ref{eq:rdpc}.  We then approximate the positions of the two-bounce spots by interpolating from the positions of the nearest (in apparent angle) one- and three-bounce spots.

Once we've computed these approximate positions we can estimate the mirrored source position $L' = [x_0, y_0, z_0]^T$ by solving the following optimization problem

\begin{equation}
\label{eq:opt}
	\{ x_0, y_0, z_0 \} = \mathrm{arg}\min_{x_0, y_0, z_0} \frac{1}{2} \sum_i \left[ (x_i - x_0)^2 + (y_i - y_0)^2 + (z_i - z_0)^2 - (ct_i - r_i)^2\right]^2 .
\end{equation}

Here $(x_i, y_i, z_i)$ is the approximate position (in $C$-centered coordinates) of the $i^{th}$ detected two-bounce spot, $r_i = \sqrt{x_i^2 + y_i^2 + z_i^2}$ and $t_i$ is the two-bounce travel-time associated with the $i^{th}$ spot.  The objective function is twice-differentiable, so we can solve the problem using Newton's method.  In practice, misclassifications of two-bounce spots as one- or three-bounce spots, or vice-versa, produce large errors in the source localization result.  To make our algorithm more robust to misclassification errors, we solve Eq. \ref{eq:opt} using a RANSAC \cite{fischler1981} approach that is robust to outliers.  


\subsubsection{Determining Scattering Points}  As can be seen in Figure \ref{fig:winMB}(b), rays drawn from the camera and from the true source to two-bounce points lying \emph{behind} the mirror plane will intersect the mirror plane at reflection points, as will rays drawn from the mirrored source to two-bounce spots \emph{in front of} the mirror plane.  Thus, once the mirror plane has been computed from our estimate of $L'$, we can find all points of specular reflection.  From these reflection points we can also estimate the mirror's boundary.  This allows us to determine which spots correspond to three-bounce paths.  The apparent positions of three-bounce spots can then be reflected across the mirror plane to retrieve the true diffuse scattering positions.  This can be achieved even when the true scattering points lie outside of the receiver's field-of-view (e.g. if they are hidden around a corner).

It is important to note that Eq. \ref{eq:opt} only applies to planar specular surfaces, and so this algorithm will not produce accurate reconstructions of non-planar specular surfaces. 





\section{Results}

\subsection{Implementation}

\begin{figure}[!t]
\centering
    \includegraphics[width=0.8\textwidth]{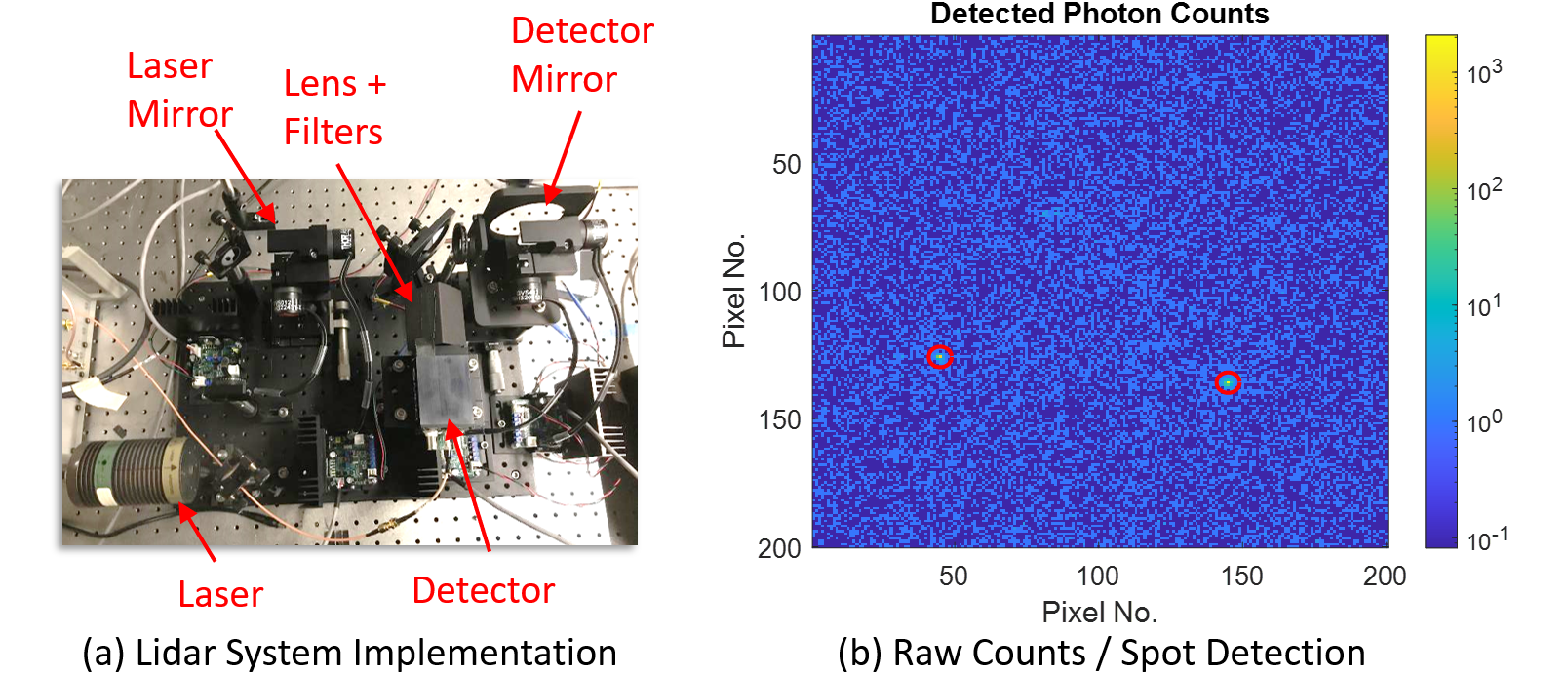}
    \caption{\textbf{Implementation}. (a) Our lidar system consists of a focused, pulsed galvo-scanned laser and a single-pixel SPAD that can be scanned independently of the laser.  In our experiments the SPAD FOV is scanned across a dense, uniform grid to mimic the angular sampling pattern of a SPAD array camera. (b) Per-pixel map of detected photon counts.  Detected spots are circled in red.}
    \label{fig:imp}
\end{figure}

\paragraph{Lidar System.}  A photo of our lidar system is provided in Fig. \ref{fig:imp}(a).  The transmitter consists of a focused, pulsed laser source (640nm wavelength) that is scanned using a two-axis mirror galvanometer.  The receiver is a single-pixel SPAD detector with a focused FOV that can be scanned independently from the laser using a second set of galvo mirrors.  The overall instrument response function (IRF) of our system was measured to be 128 ps (FWHM).  Details concerning the specific equipment used can be found in the supplemental material.


For each beam pointing direction, we reproduce the angular sampling pattern of a SPAD array camera by scanning the FOV of the detector across a dense, uniform grid.  For experiments reported in this paper, the per-pixel dwell time was either 5 or 10 ms.  The laser was operated at a pulse repetition frequency of 20 MHz and an average transmitted power of 5 $\mu W$.

\paragraph{Spot Extraction.}  Raw photon count measurements are sorted into a data cube that is binned by photon time-of-arrival and detector scan angle.  From these raw measurements we detect spot-like returns and then extract the time-of-flight, angle-of-arrival, and returned energy of each spot.  An image of per-pixel detected photon counts collected from a single detector scan is shown in Fig. \ref{fig:imp}(b).  Detected spots are circled in red.  A more detailed description of our spot detection and low-level signal processing pipelines is provided in the Supplement.

\subsection{Planar Surface Scans}
\label{sec:planar_results} 

\begin{figure}[!t]
\centering
    \includegraphics[width=\textwidth]{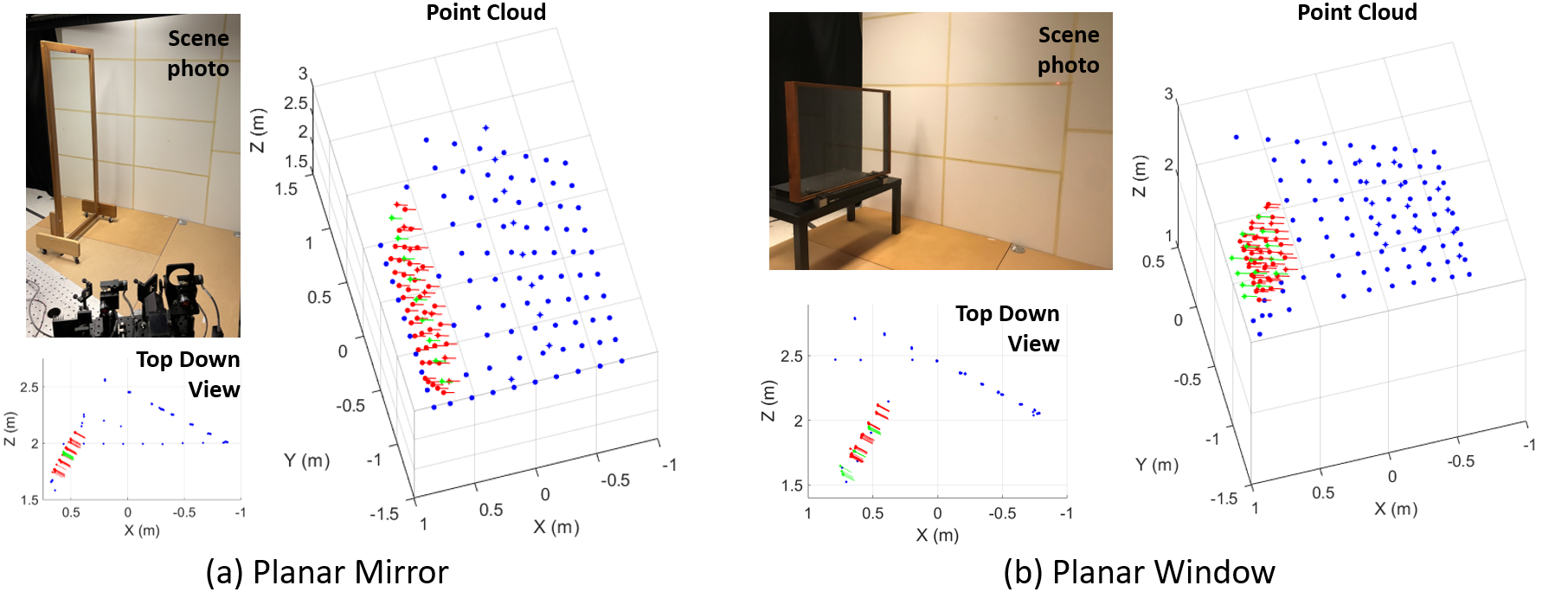}
    \caption{Scans of (a) a large flat mirror and (b) a single-pane glass window.  Red and green points indicate points on specular surfaces, blue points indicate points on diffusely reflecting surfaces.  Circles indicate diffuse-first illumination geometry and asterisks indicate specular-first geometry.  Surface normals are also plotted at specular points.}
    \label{fig:mirror_res}
\end{figure}


\subsubsection{Planar Mirror}  
We scanned a scene (shown in Figure \ref{fig:mirror_res}(a)) that contained a tall, flat mirror, a wooden floor, and a matte white wall.  The mirror was placed approximately 2m away from the lidar scanner.  We scanned the laser beam along a $10\times10$ grid of pointing directions.  14 beams illuminated the mirror directly.  The remainder illuminated the floor, wall, or the mirror's wooden frame.  For each beam direction the detector FOV was scanned across a $200\times200$ grid that subtended a $\pm 30^\circ$ angle of view in both vertical and horizontal directions.  The per-pixel dwell time was 5 ms.  Total acquisition time was 5 hours and 33 minutes, although we note that an equivalent data collection made using a $200\times200$ pixel SPAD array would have been captured in 0.5 seconds.



Our results are shown in Fig. \ref{fig:mirror_res}(a).  Here, blue points represent points on diffuse surfaces, red points are points on the mirror as seen from the receiver ($S$ or $S_2$ in Fig. \ref{fig:mirror_geom}), and green points are points on the mirror that were illuminated directly ($S_1$ in Fig. \ref{fig:mirror_geom}).  Points computed using diffuse-first equations are marked by circles, whereas points computed using specular-first equations are marked with asterisks.  Surface normals are also plotted for specular surface points.

It is evident from Fig. \ref{fig:mirror_res}(a) that the point cloud accurately captures the dimensions of the scene.  From a comparison to a ground truth scan collected on the mirror's frame, we determined that the points on the mirror surface had an RMS displacement of $9.5 mm$ with respect to the ground truth plane, and the surface normals had an RMS tilt of $0.65^\circ$ with respect to the true surface normal.  A complete evaluation of the scan's accuracy is provided in the Supplement.


\subsubsection{Planar Glass Window} We also scanned the shape of a single-pane glass window, which is shown in Fig. \ref{fig:mirror_res}(b).  The window was placed in the same position and orientation as the previously scanned mirror.  For this collection we did \emph{not} place any objects behind the window, and so the primary challenge was that the lower reflectance of the glass resulted in fainter multibounce returns, particularly for three-bounce light.  We found that each reflection off the window reduced the range-adjusted intensity of a spot by approximately a factor of 10.  



We compensated for the lower reflectance by doubling the per-pixel dwell time, to 10 ms.  The laser scan pattern was changed to an $11\times9$ grid that more densely sampled the portion of the scene that contained the window.  The detector scan pattern was identical to that used in the mirror experiment.  The total acquisition time was 11 hours, although we note that an equivalent data collection could have been captured by a SPAD array in just under one second.


Our results are shown in Fig. \ref{fig:mirror_res}.  The window pane can be seen clearly, and appears to have been mapped accurately.  Despite the lower reflectance of the glass, we are able to detect all two-bounce and three-bounce returns with no misses or false alarms  However, when the point cloud is viewed from the top-down it is clear that the points computed from three-bounce returns (marked by red and green asterisks) are noisier than two-bounce points.  This was likely a consequence of the lower relative intensity of three-bounce returns. 


\subsubsection{Detecting Objects Behind a Window} We placed two objects behind the window and the repeated the window scan described previously.  Our results are shown in Fig. \ref{fig:teaser}(a).  Behind-window detections are marked by white dots circled in blue.  From these results we see that our disambiguation logic accurately classified four spots that scattered off of the objects behind the window.  We note that all other points that appear to lie behind the window in the top-down view of the point cloud in fact lie above or below the window aperture.



In addition to the results shown here, we performed a second experiment that was designed to rigorously test the disambiguation criteria described in Section \ref{sec:window}.  These additional results can be found in the supplemental material in Section S.4.

\paragraph{Comparison to Na\"ive Single-bounce Processing.}  On the right side of Fig. \ref{fig:teaser}(a) we show the point cloud that would have been generated if the positions of all spots had been na\"ively computed using the conventional one-bounce range equation.  Notably, this na\"ive procesing detects \emph{no} points on the window's surface.  Detections that might have been mapped to the window surface are instead mapped to erroneous points.  Only true one-bounce returns are mapped correctly.   

\subsection{Non-planar Object Scans}
\label{sec:nonplanar_results}

Our method can also be used to acquire the shape of non-planar specular surfaces.  We demonstrated this by scanning the shape of a copper pitcher.  A photo of this pitcher and the results of our scan are shown in Fig. \ref{fig:teaser}(b).   The pitcher was placed on a wooden floor and in between two white side walls.  We illuminated a sequence of 60 laser spots on these surfaces, and observed two-bounce returns that reflected off of the pitcher.


Laser spots and two-bounce highlights were acquired using a $200\times100$ pixel scan with a 5 ms per-pixel dwell time.  We repeated the scan four times, rotating the pitcher $90^\circ$ between each scan so that we could acquire it's front, back, and side-facing surfaces.  The four point clouds were aligned to a pitcher-centered coordinate system and then combined.  From the combined point cloud in Fig. \ref{fig:teaser}(b), we see that were are able to recover the general shape of the pitcher, although there are several gaps that correspond to regions that did not produce any detectable highlights due to their surface orientation.

We point out two distinctive features of the non-planar surface scanning process.  First, because the pitcher's surface alternated between convex, concave, and hyperbolic curvature, one laser spot would produce multiple highlights, each of which could be used to locate a point on the pitcher's surface.  This can be seen in the photograph on the top-right of Fig. \ref{fig:teaser}.  Second, Eqs. \ref{eq:rdc3b} and \ref{eq:rs2c} are only correct when the tangent planes of surface points $S_1$ and $S_2$ are coincident.  This is automatically satisfied for monostatic ($\mathcal{s}=0$) lidar systems, for which $S_1=S_2$.  However, because our scanner was bi-static, these equations could not be applied for non-planar surface mapping.  Consequently, all points in Fig. \ref{fig:teaser} had to be acquired by directly illuminating spots on diffuse reflectors, and observing two-bounce returns reflected by the pitcher.  Specular-first returns could not be used.  However, using the criteria outlined in Sec. \ref{sec:geom}, specular-first returns could be detected automatically and discarded.


\subsection{Multiple-Beam Illumination}

\begin{figure}[!t]
\centering
    \includegraphics[width=\textwidth]{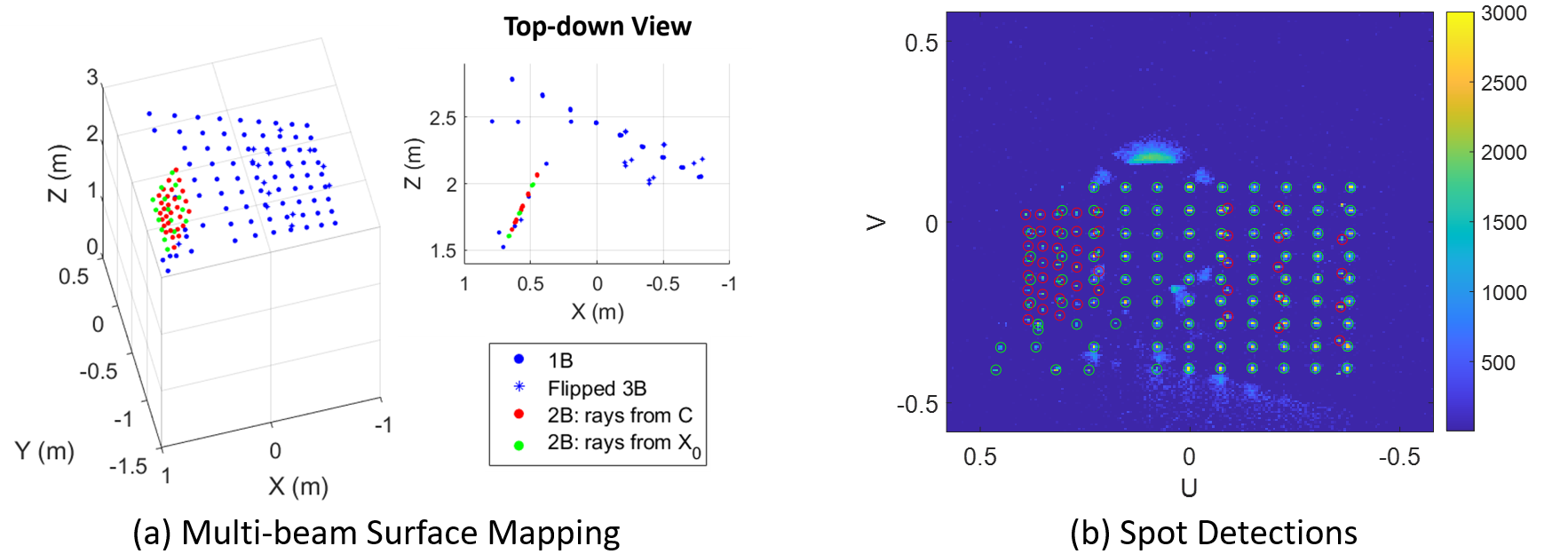}
    \caption{(a) Point-cloud generated using multi-beam surface mapping algorithm on glass window dataset (Fig. \ref{fig:mirror_res}b). (b) Detected per-pixel energy values of multi-beam data collection, overlaid with spot positions.  Red circles mark two-bounce spots, and green circles mark one- or three-bounce spots.}
    \label{fig:mb_res}
\end{figure}

To test our multi-beam surface mapping algorithm, we summed the single-beam photon count histograms acquired during the first planar window scan described in Sec. \ref{sec:planar_results}.  This produced the equivalent of a single multi-beam flash exposure.  Detected per-pixel energy values from this multi-beam data are shown in Fig. \ref{fig:mb_res}(b).  One- and three-bounce spots are circled in green, and two-bounce spots are circled in red.
%

Our results are shown in Fig. \ref{fig:mb_res}(a).  By visual inspection we see that the window scattering points closely match the single-beam results in Fig. \ref{fig:mirror_res}(b).  Points in the multi-beam results in fact appear less noisy, although this is because they are  intersections with a single, fitted plane.  Many reflected three-bounce points appear in erroneous positions.  Most of these points correspond to two-bounce spots that were incorrectly classified as three-bounce spots. 



\section{Discussion}

We have demonstrated methods that use multibounce specular lidar returns to detect and map specular surfaces that might otherwise be invisible to lidar systems that rely on single-scatter measurements.  We considered the cases of single-beam and multiple-beam illumination, and demonstrated our methods by scanning planar, non-planar, and transparent surfaces.

Because specular surfaces are relatively common, our work could be used in most domains that lidar scanning is applied to, including autonomous navigation, mapping of indoor or outdoor spaces, and object scanning.  Future work might address failure cases, such as propagation paths with consecutive specular reflections, or geometries for which the true or mirror image spot lies outside the receiver FOV.  We would also be interested in extending our multi-beam algorithm to the mapping of curved surfaces, and to allow dense flash illumination.

\section{Acknowledgements}  Connor Henley was supported by a Draper Scholarship. This material is based upon work supported by the Office of Naval Research under Contract No. N00014-21-C-1040.  Any opinions, findings and conclusions or recommendations expressed in this material are those of the author(s) and do not necessarily reflect the views of the Office of Naval Research.


\clearpage
%
%
\bibliographystyle{splncs04}
\bibliography{references}
\end{document}


\pagestyle{headings}
\mainmatter
\def\ECCVSubNumber{7438}  

\title{Detection and Mapping of Specular Surfaces Using Multibounce Lidar Returns} \subtitle{Supplementary Information}

\titlerunning{Detection and Mapping of Specular Surfaces}
%
\author{Connor Henley \inst{1,2} \and
Siddharth Somasundaram\inst{1} \and
Joseph Hollmann \inst{2} \and
Ramesh Raskar \inst{1}}
\authorrunning{C. Henley et al.}
%

\institute{Massachusetts Institute of Technology\\
\email{\{co24401,sidsoma,raskar\}@mit.edu}\\
\and
The Charles Stark Draper Laboratory\\
\email{jhollmann@draper.com}}
\maketitle

\renewcommand{\thefigure}{S.\arabic{figure}}
\renewcommand{\theequation}{S.\arabic{equation}}
\renewcommand{\thesection}{S.\arabic{section}}

\section{Summary of Multi-beam Surface Mapping Algorithm}
Our algorithm can be summarized by the following steps:

\begin{enumerate}
    \item Intersect beam vectors with apparent spot positions to determine one-bounce (1B) and three-bounce (3B) spots.  Classify other spots as two-bounce (2B).
    \item Compute apparent positions of 1B and 3B spots using Eq. 1. 
    \item Approximate 2B spot positions by interpolating from nearest 1B and 3B spot positions.
    \item Use 2B times-of-flight and approximate positions to solve mirrored source localization problem (Eq. 8). 
    \item Compute mirror plane parameters from mirrored source position.
    \item Retrieve scattering points on mirror surface by drawing lines to 2B spots from camera and from mirrored source.  Scattering points are points of intersection with mirror plane.
    \item Compute angular positions (from camera perspective) of mirror scattering points.  Classify non-2B spots that appear within convex hull of these angular positions as 3B spots.
    \item Reflect apparent positions of 3B spots (computed in step 2) across mirror plane to retrieve true diffuse scattering points.
\end{enumerate}

It is important to note that Eq. 8 only applies to planar specular surfaces, and so this algorithm will not produce accurate reconstructions of non-planar specular surfaces.  We believe that Eq. 8 could be modified to support surface of constant curvature, however we do not attempt to do so in this work.

\subsection{Robust Source Localization}
\label{sec:supp_ransac}

In practice, misclassifications in step 1 can result in large errors in the source location computed in step 4.  To make our algorithm more robust to misclassification errors, we compute a solution to Eq. 8 using RANSAC \cite{fischler1981}, which is robust to outliers.  In our implementation, the parameters of the fitted model are the coordinates $\{x_0, y_0, z_0\}$ of the mirrored source position.  At each iteration we produce tentative source position fits to sets of $n$ points, and find inliers to this fit among the set of all other points.  If there are greater than $d$ additional inliers, we incorporate these inliers into an improved fit.  We run the RANSAC procedure for $k$ iterations, and the best model is selected to be the model with the lowest mean squared error among it's inlier set.  For results reported in this paper, we used values of $n=4$, $d=10$, and $k=1000$.

\section{Implementation: Additional Details}


\paragraph{Lidar System} A photo of our lidar system is provided in Fig. 4(a) of the main text.  The transmitter consists of a focused, pulsed laser source (Picoquant LDH-D Series) with a wavelength of 640 nm that is scanned using a two-axis mirror galvanometer (Thorlabs GVS012).  The receiver is a single-pixel SPAD detector (MPD PDM series) with a focused FOV that can be scanned independently from the laser using a second set of galvo mirrors (Thorlabs GVS412).A 50 mm lens was used to focus the SPAD FOV at infinity, and a bandpass filter centered at 640 nm was placed in front of the lens to minimize ambient background interference.  A Picoquant Hydraharp 400 was used to synchrnously record photon arrival times.  The timing jitter of our SPAD was 50 ps (FWHM) and the pulsewidth of our laser was 90 ps (FWHM).   The overall instrument response function (IRF) of our system was measured to be 128 ps (FWHM).

\paragraph{Low-level Signal Processing.}  Raw photon count measurements are sorted into a data cube that contains a time-of-flight histogram for each angle, or pixel, of the detector scan.  Each histogram is convolved with a matched filter that resembles the system IRF.  A detection is declared if the matched peak is greater than a noise-dependent false-alarm threshold and a hand-tuned absolute counts threshold.  A fitting procedure (details are provided in \cite{BFTCI}) estimates the return pulse's time-of-arrival in each pixel to sub-timing-bin resolution using counts within a small fitting window surrounding the matched peak.  The estimated uncertainty in each time-of-flight estimate is also computed.  The return energy in each pixel is reported as the sum of the photon counts within this fitting window minus the expected number of background counts.  Counts in a designated set of noise bins are used to estimate the background level in each histogram.

\paragraph{Spot Extraction} Per-pixel energy values, time-of-flight values, time-of-flight uncertainties, and boolean detection flags are passed to a spot extraction pipeline.  The energy image is then filtered by a matched ``spot" filter (which is a laplace filter when a $3\times3$ filter size is used) and a box filter of the same size.  The ratio of these filter outputs is computed at each pixel as a measure of ``spottiness".  A ``spottiness" threshold is employed to filter out intense but un-spot-like returns, such as the diffuse inter-reflections seen near edges and corners.  We find contiguous regions of pixels that have passed both the spottiness test and the low-level detection criteria described previously.  In each of these regions we find the pixel with the maximum energy value and center a small ($3\times3$ or $5\times5$) window around that pixel.  The spot center is set to be the centroid of the energy values within the window.  This spot center is first computed in pixel space, and then converted to angle-space to retrieve the angle-of-arrival of the spot.  The spot's energy is set to the sum of the energy values within the window, and the spot's time-of-flight is set to be the uncertainty-weighted average of the time-of-flight values within the window.  If two spots are too close together (less than half the window size, rounded up), we throw away the spot that has the lower energy.


\section{Analysis of Errors in Mirror Scan Experiment}

To validate our method, we investigated the accuracy of the points computed for our scan of the large mirror that was reported in Sec. 4.2 of the main text.  We fit a plane to a set of 22 ground truth points on the mirror's frame.  These ground truth points were acquired during a separate data collection which utilized a dwell time of $25 ms$ per pixel---which was five times higher than the dwell time of the original scan.  This longer dwell time allowed us to collect more photons per point and thus make more precise position estimates.  All points on the mirror frame reflected light diffusely, so their positions could be computed from the conventional single-scatter range equation.  The ground truth plane fit was then adjusted to account for the fact that the mirror surface was recessed $1.6cm$ behind the plane of the frame.

\begin{figure}[!t]
\centering
    \includegraphics[width=\textwidth]{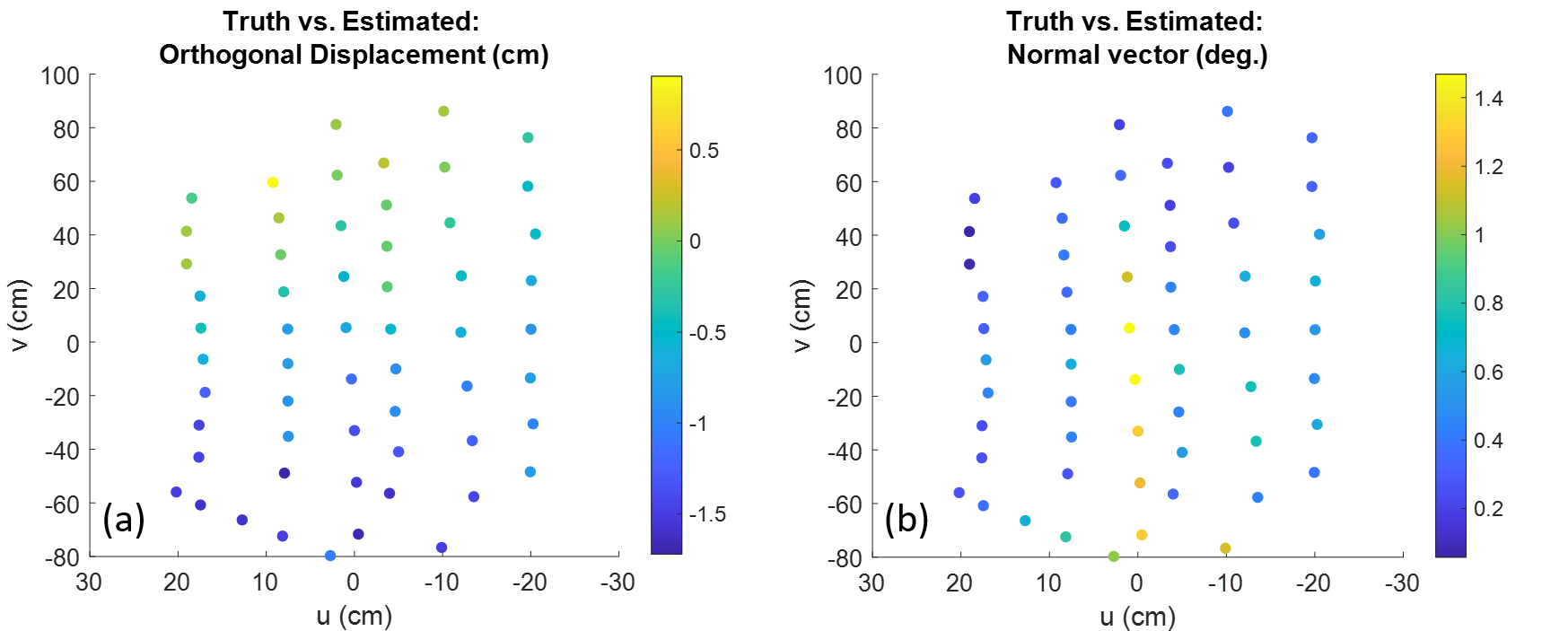}
    \caption{\textbf{Mirror scan error analysis.}  Plots of (a) distance of points from ground truth mirror plane and (b) angular displacement of surface normal vectors from the ground truth plane.  Errors are plotted as a function of the projected position of each point onto the ground truth plane.}
    \label{fig:errors}
\end{figure}

We then compared the position and surface normal of mirror surface points computed from the original scan data to this ground truth plane.  In Figure \ref{fig:errors}(a) we plot the orthogonal displacement of all estimated mirror plane points as a function of their projected position on the ground truth plane.  The root-mean-square (RMS) displacement of these points was $9.4 mm$ and the mean displacement was $-7.1 mm$.  In Figure \ref{fig:errors}(b) we show the angular displacement of the estimated normal vectors from the ground truth normal vector.  The RMS angular displacement was $0.63^\circ$ and the mean displacement was $0.54^\circ$.  The presence of small but significant mean errors suggest a systematic bias in our results.  This bias could have resulted from many factors.  First, the angular calibration of our system was not perfect, and errors in angle-of-arrival estimates of diffuse surface returns may have been coupled into specular surface positions.  Second, the ground truth scan was taken 24 hrs after the initial scan, and it is possible that the mirror moved or tilted slightly in between scans.  Finally, the presence of a glass first-surface may have also contributed to the bias.  The mirror's protective glass was $6.35 mm$ thick, and we did not account for the reduced speed of light within the glass while estimating the positions of points on the mirror surface.  To do so would have been non-trivial because the increase in travel time would have been dependent on angles of incidence and exittance with respect to the glass surface.  Another prominent artifact is the larger normal vector errors associated points computed from specular first-surface measurements as opposed to diffuse first-surface measurements.  These points form a vertical line down the center of the plane which can be seen plainly in Figure \ref{fig:errors}(b).

\begin{figure}[!t]
\centering
    \includegraphics[width=\textwidth]{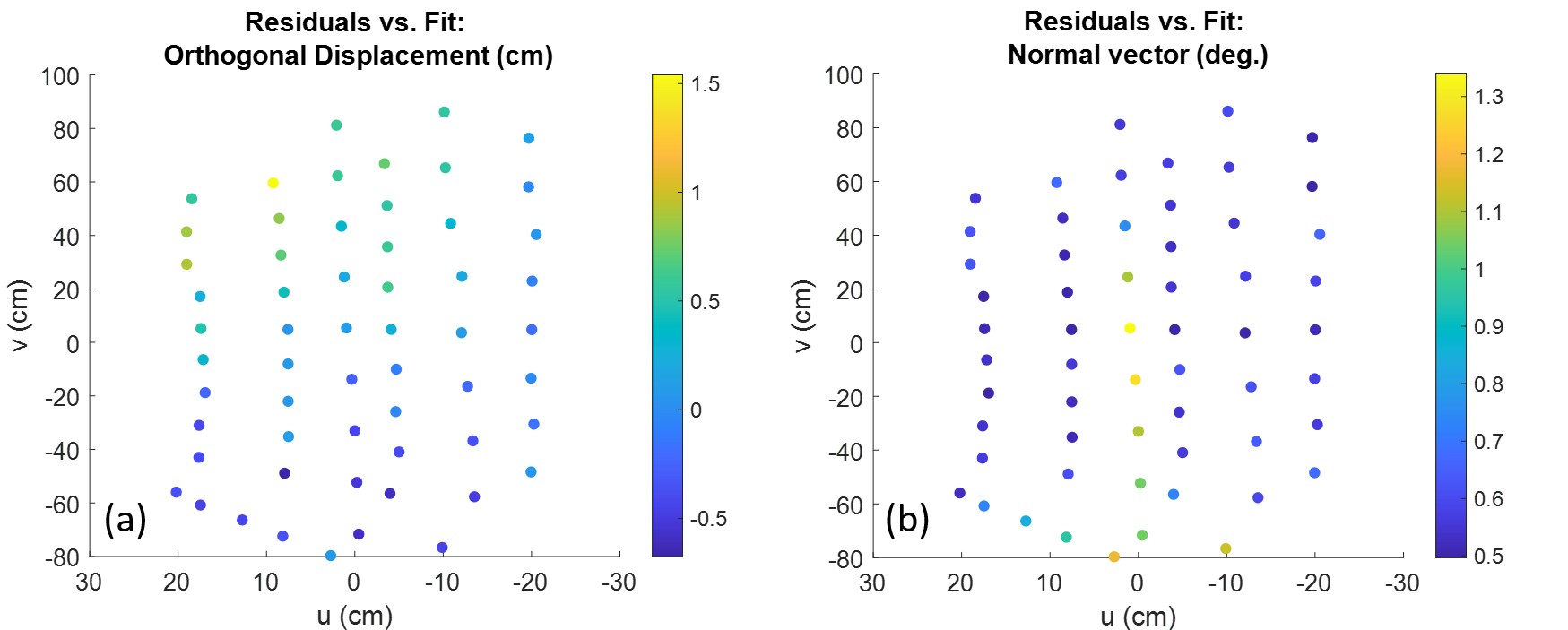}
    \caption{\textbf{Mirror scan residuals analysis.}  Plots of (a) distance of points from the fitted mirror plane and (b) angular displacement of surface normal vectors from the fitted plane.  Residuals are plotted as a function of the projected position of each point onto the fitted plane.}
    \label{fig:resids}
\end{figure}

To assess the precision of our estimates, we fit a plane to the collection of mirror surface points and then compared individual points to the plane fit.  Fitting a plane to oriented points is straightforward.  Because each oriented point includes a position and a surface normal vector, each oriented point defines a plane.  Thus, to ``fit" a plane to these oriented points, we simply need to compute the plane parameters for each individual point and then take the average of these parameters.  We define a plane by the identity $\hat{n}\mathbf{x} = d$.  The fit plane had the parameters $\hat{n}_{fit} = [-.8797 -.0048 -.4754]^T$ and $d_{fit} = -1.406 m$.  For comparison, the ground truth plane had parameters  $\hat{n}_{truth} = [-.8825 -.0010 -.4704]^T$ and $d_{truth} = -1.389 m$.  We plot the orthogonal displacement of all estimated mirror plane points with respect to the fitted plane in Figure \ref{fig:resids}(a) .  We show the angular displacement of the estimated normal vectors from the mean normal vector in Figure \ref{fig:resids}(b).  The RMS orthogonal distance was $4.7 mm$ and the RMS  angular displacement was $0.70^\circ$.

\section{Additional Experiment: Transparent Specular Surface}

\begin{figure}[!t]
\centering
    \includegraphics[width=\textwidth]{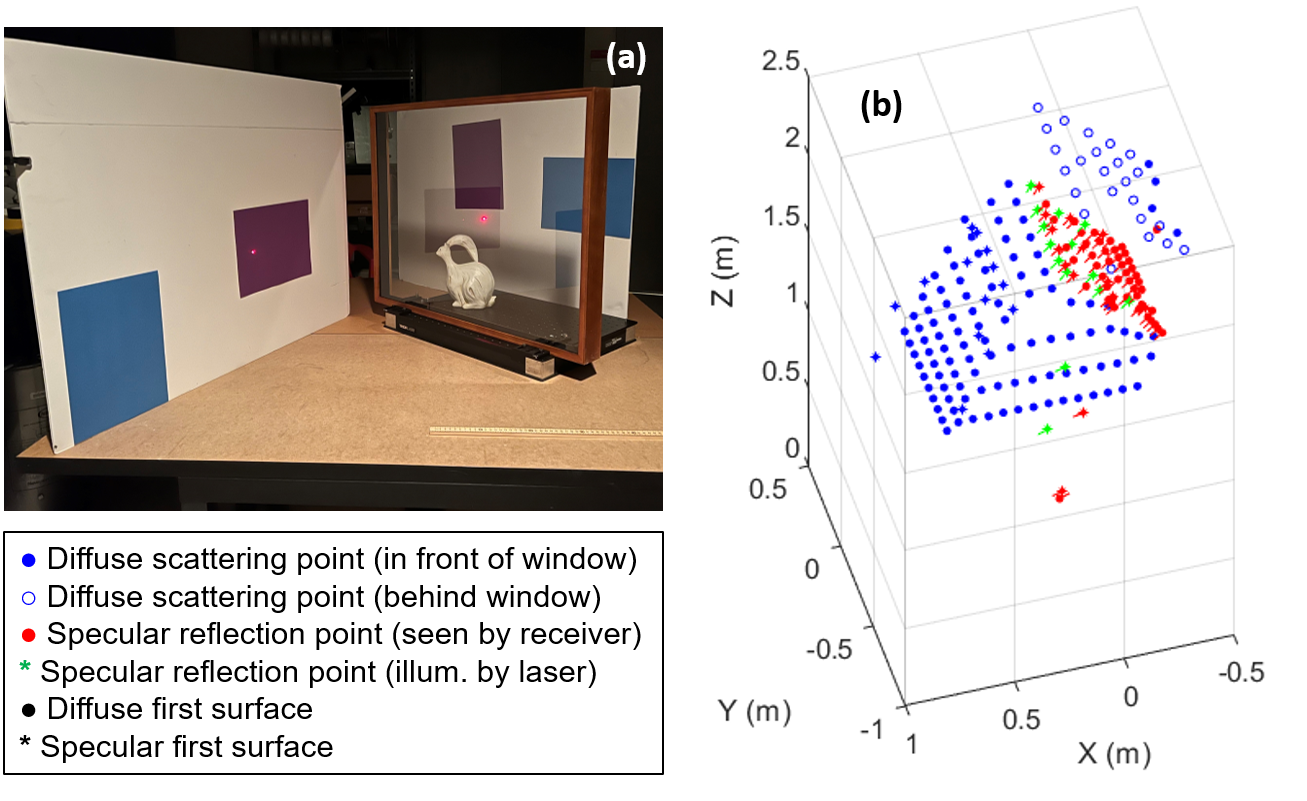}
    \caption{(a) Photograph of scanned scene which included a glass window as well as surfaces in front of and behind the windw that had varying reflectance. (b) Point cloud acquired for the scene.  Marker key is shown on the bottom left.}
    \label{fig:window_disambig}
\end{figure}

We carried out an additional experiment to more exhaustively test the transparent specular surface processing logic that was proposed in Sec. 3.2 of the main text.  In this experiment we scanned a scene that contained two large, diffusely reflecting walls that were respectively placed in front of and behind a single-pane glass window.  A photograph of this scene is shown in Fig. \ref{fig:window_disambig}(a).  Dark pieces of construction paper were pasted onto each wall to produce surfaces that varied between high and low reflectance.  At some view angles the mirror image of the front wall appeared to be closer to the camera than the true plane of the back wall, whereas at other view angles it appeared to be more distant.   This assured that both the time-of-flight and reflectance criteria proposed in Sec. 3.2 were put to the test. For added scene complexity, we also placed a small rabbit figurine immediately behind the window.

We scanned the scene using a pattern of $14\times 10$ beams scanned in a uniformly spaced rectangular grid.  The detector scan was $190\times140$ pixels that spanned $50.3^\circ\times37.3^\circ$ with a per-pixel dwell time of $10 ms$.  Total scan time was 11 hours, 7 minutes, and 20 seconds, although a SPAD array could have made an equivalent measurement in 1.4 seconds.

\subsubsection{Undetected Three-bounce Highlights} In contrast to the experiments described in the main text, in this scene there were many occasions where the three-bounce reflection off of the window surface was too faint to detect.  This frequently occurred when the laser beam was deflected onto a dark surface on the front wall.  Losses from absorption at the diffuse surface compounded with losses from two glass reflections, resulting in extremely faint three-bounce returns that could not be detected.  Our original processing logic could not be used in this scenario, because it relied on the presence of multiple detections along the beam vector to detect the presence of a window.  

We addressed this scenario by implementing an additional test.  As in our regular pipeline, if only two spots were detected, we assumed that the spot that was associated with the lower time-of-flight measurement corresponded to a true spot on a diffuse surface lying in front of the window ($D_1$ in Fig. 3(a)).  If the range-adjusted reflectance (computed using intensity and na\"ive single-scatter range) of this first spot was \emph{lower} than the range-adjusted reflectance of the second spot, we infered that the second spot was in fact a diffuse return from a surface behind the window ($D_2$ in Fig. 3(a)).  The reasoning behind this is that a true three-bounce return should always have a lower apparent reflectance than the two-bounce return that it is associated with, due to the additional reflection off of the glass surface.

This line of reasoning fails in two important cases.  First, if the time-of-flight associated with the behind-window detection is smaller than the time-of-flight associated with the spot in front of the window, then the true spot in front of the window will be interpreted as a return from a specular surface ($S$ in Fig. 2(a)).  Second, if the illuminated spot behind the window has very low reflectance, it might be interpreted as a three-bounce highlight.  However, there is no way to directly disambiguate these scenarios using only the relative times-of-flight and reflectance measurements.

\subsubsection{Results.}

Our results are shown in Fig. \ref{fig:window_disambig}(b).  We are able to retrieve the surface of the window as well as the surface of the two large walls and the floor.  A single detection off of the rabbit figurine was also registered.  However there are a few artifacts in our point cloud that are worth pointing out.  

First, three (of 140) beams produced detections in false positions.  These detections can be seen floating in the space  between the lidar system and the window, and floating behind the front wall.  In all three cases, the true three-bounce returns were not detected.    Two of these beams illuminated a spot on a dark surface lying behind the window.  The behind-window spot was misinterpreted as a three-bounce return, and the wrong set of range equations was applied, resulting in three erroneously mapped detections in each case.  For the third beam, the time-of-flight of the behind-window spot was smaller than the time-of-flight arriving from the spot in front of the window.  This meant that the spot on the front wall was interpreted as a specular reflection point, and was erroneously mapped to a point directly in front of the receiver.

Second, many points on the back wall are missing or misclassified as lying in front of the window.  Most of the missing points correspond to portions of the wall that lay in the shadow of the rabbit figurine.  The remaining missing points correspond to the misinterpreted behind-window spots that were described in the previous paragraph.  Four spots lying behind the window were classified as lying in front of the window.  This occurred during exposures for which the spot behind the window was the only spot detected, and was thus treated as a regular spot on a diffusely reflecting surface.


\section{Multi-beam Scan of Large Mirror Target}

\begin{figure}[!t]
\centering
    \includegraphics[width=\textwidth]{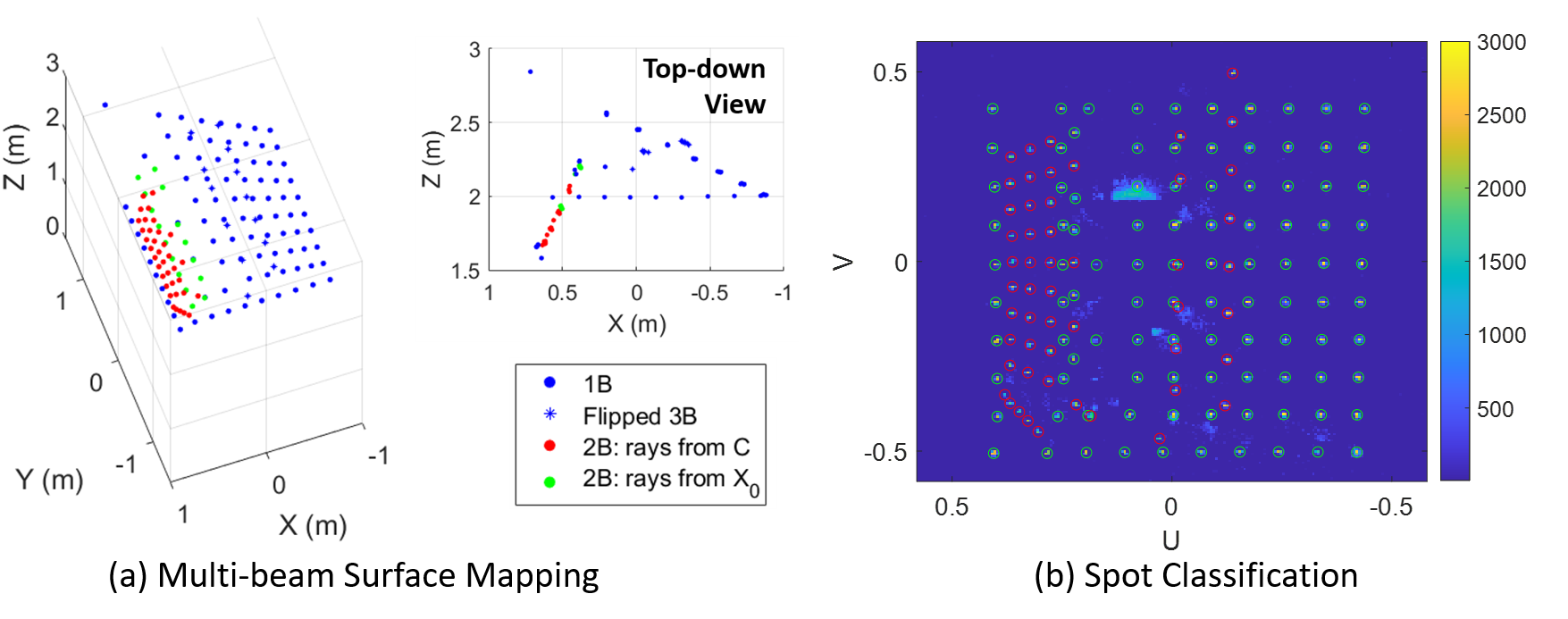}
    \caption{(a) Point-cloud generated using multi-beam surface mapping algorithm on large mirror dataset (Fig. 4a). (b) Detected per-pixel energy values of multi-beam data collection, overlaid with spot positions.  Red circles mark spots classified as two-bounce, and green circles mark spots classified as one- or three-bounce.}
    \label{fig:mb_mirror}
\end{figure}

We also tested our multi-beam surface scanning algorithm on the data collected for the large mirror scan reported in Sec. 4.2 of the main text.  As with our multi-beam scan of the window, we summed together the photon-count histograms acquired from the single-beam scan of the mirror to generate the equivalent of one multi-beam flash exposure.  Detected per-pixel energy values of this summed dataset are shown in Figure \ref{fig:mb_mirror}(b).  Spots were classified by intersecting naive position estimates of all spots with the set of all transmitted beam vectors.  Spot positions that lay within some angular distance of a beam vectors were classified as potential one- or three-bounce spots (circled in green in Fig. \ref{fig:mb_mirror}(b)).  The remainder were classified as two-bounce spots (circled in red in Fig. \ref{fig:mb_mirror}(b)).  If multiple spot positions were deemed to intersect the same beam vector, the spot closest to the beam vector was chosen as the candidate one- or three-bounce spot, and the others were classified as two-bounce spots.  This classification process was not perfect, and misclassification errors justified our use of a robust RANSAC \cite{fischler1981} source fitting algorithm that was described in Sec. \ref{sec:supp_ransac}.

Our results are shown in Fig \ref{fig:mb_mirror}(a).  As with the window scan, the results of the multi-beam surface scanning algorithm produced for the large mirror target match the single-beam results very closely.  Interestingly, because our multi-beam algorithm did not require spot-to-beam associations, our multi-beam algorithm was able to reclaim certain points that were not computed during the original single-beam scan.  These additional points corresponded to points directly illuminated near the edge of the mirror's surface ($S_1$ in Fig. 2(b)).  In the original scan, no highlight ($S_2$) was observed for these beams because the point of reflection would have been beyond the mirror's edge, and thus $S_1$ was not computed.  However, in the multi-beam algorithm, the intersections of all beam vectors with the fitted mirror plane are computed, including the points that had no associated highlight measurement.

\section{Scan of a Thin Transparent Object}

\begin{figure}[!t]
\centering
    \includegraphics[width=0.9\textwidth]{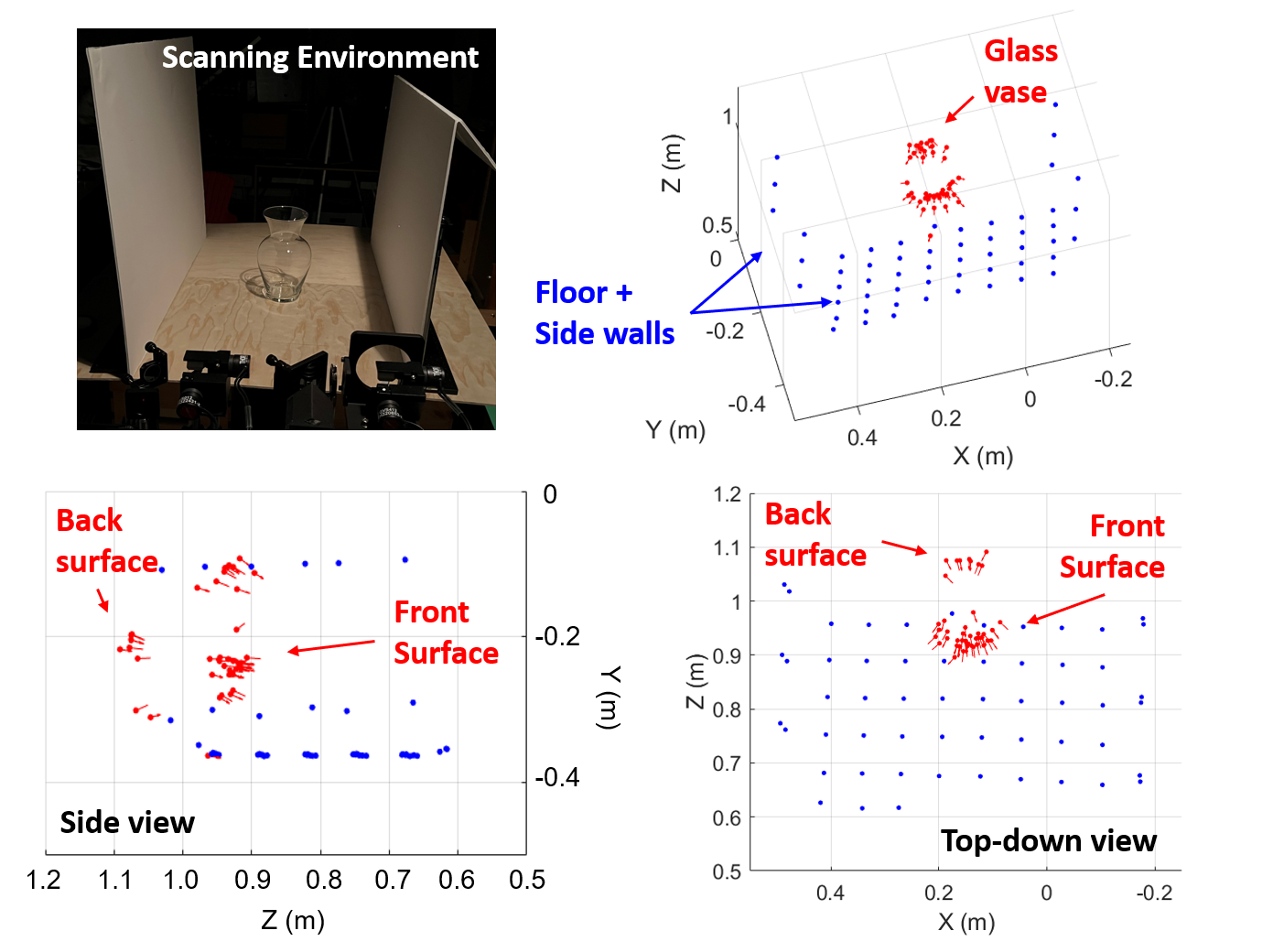}
    \caption{We scanned the shape of a glass vase (top left) by directly illuminating a sequence of 60 laser spots on nearby surfaces and measuring the time-of-flight of two-bounce returns that reflect off of the pitcher.   We show the estimated point cloud from multiple views (top right and bottom row).  Blue dots are points on the floor and side-walls, red dots are points on the vase, and red arrows are the surface normal vectors associated with those points.  Because the vase was transparent, we were able to map points on the front and back surfaces of the vase without moving our lidar, or rotating the vase. }
    \label{fig:vase}
\end{figure}

In addition to scanning a metal pitcher, we also scanned a glass vase.  We used the same detector and laser scan patterns that were used to acquire the shape of the copper pitcher in Sec. 4.3 of the main text, but increased the per-pixel dwell time of our detector to $10 ms$ to account for the lower reflectivity of the glass.  

Our result is shown in Fig. \ref{fig:vase}.  Due to the rotational symmetry of the vase, we did not rotate it to measure the shape of all sides.  Interestingly, after scanning the vase from just a single vantage point, we noticed that we were able to measure points on both the front \emph{and} the back surface of the vase, due to the fact that the vase was transparent.  Furthermore, because the vase is a \emph{thin} glass object, we expect that the effect of distortion of the returned signals due to refraction should be small.  Thus, the position of these back surface points should be relatively accurate.  Surface normals for back-surface points are identified as inward-facing, whereas surface normals on the front surface are outward facing.  This is technically correct, and in each case corresponds to the true direction of the surface normal vector at the point of specular reflection.  

\section{Common Failure Cases}

\begin{figure}[!t]
\centering
    \includegraphics[width=\textwidth]{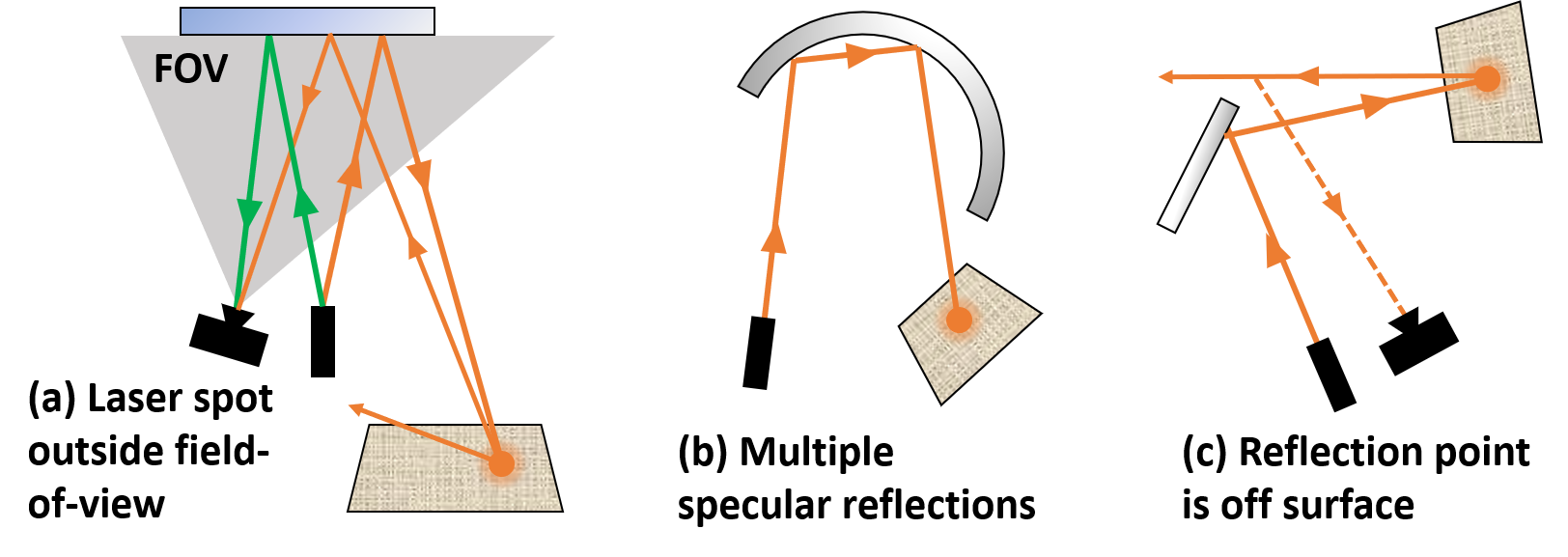}
    \caption{\textbf{Common Failure Cases.}}
    \label{fig:fail}
\end{figure}

Our methods can be used to detect and localize specular surfaces that might be invisible to a conventional lidar system that relies on measurements of direct, single-scatter returns.  Nevertheless there are specific circumstances that cause our method to fail.  We depict some of these cases in Figure \ref{fig:fail}.  

\subsubsection{Only one spot is detected.}  When the scene is illuminated with a single beam, the position of the specular surface can only be determined if \emph{both} the true laser spot and its mirror image are detected.  There are many scenarios in which this condition is not met.  The most obvious is when the laser beam reflects off of a specular surface but never lands on a diffusely reflecting surface.  This scenario is more common in wide-open outdoor environments in which the beam might get reflected towards the sky.  A scenario that is more common in enclosed environments is when the beam is deflected onto a surface that lies outside the receiver's field-of-view (FOV).  We illustrate this scenario in Fig. \ref{fig:fail}(a).  Such a scenario often occurs when a specular surface is illuminated at near-normal incidence, such that the beam lands on a diffuse reflector that is behind the lidar.  One remedy in this context is to use a receiver with a very wide FOV.  Although such receivers may not be typical, many lidar systems used by autonomous vehicles have a $360^\circ$ horizontal FOV. Additionally, while near-normal incidence angles challenge methods that rely on specular multibounce returns, they are also more conducive to observations of the direct, single-scatter returns that can be interpreted using conventional processing.  Lastly, it is also possible that an occluding surface will block the receiver's line-of-sight to the true laser spot.

The consequences of detecting only one of two spots depend upon which spot is detected and which surface was illuminated first.  If a diffuse reflector is illuminated first and only one-bounce returns are visible, then the diffuse reflection point can be determined accurately using Eq. 1 but we will learn nothing about the specular reflector.  If the specular reflector is illuminated first and only the three-bounce return is observed, then the three-bounce return may be interpreted as a one-bounce return, causing a false point to be mapped behind the specular surface.  Regardless of which surface was illuminated first, if only the two-bounce return is detected then it can be identified and discarded without producing false points.  This is because the direction that two-bounce light returns from will not, in general, match the direction of the transmitted beam.


\subsubsection{Multiple specular reflections.}  Our method does not consider propagation paths that include two or more specular reflections in a row.  We've depicted one of these propagation paths in Figure \ref{fig:fail}(b).  This scenario can occur when scanning a free-form specular object with concavities, or when a scene has multiple planar reflectors that face each other.  Although we do not treat such cases in this work, it is likely that the additional specular points can be mapped in these circumstances if the additional highlights that they produce are observed.  Alternatively, certain cues such as chirality flips of an asymmetric beam pattern might be used to identify and discard multi-specular returns.

\subsubsection{Reflection point is off surface.} If there is a non-zero baseline separation between the transmitter and the receiver (that is, if the lidar system is bi-static), then there will be a disparity between the point that the transmitter directly illuminates on a specular surface ($S_1$) and the second point of reflection ($S_2$), which corresponds to the specular highlight that is visible to the receiver.  This sometimes leads to the scenario depicted in Fig. \ref{fig:fail}(c) when a specular surface is illuminated near its edge.  Here, the point that \emph{would} have reflected light back towards the receiver is just empty space---it's beyond the edge of the specular surface.  In this case no three-bounce return is observed, but a two-bounce return is still produced.  In such a scenario, $S_1$ can be recovered by finding the point at which the beam vector intersects a plane (or other shape) fit to \emph{other} specular surface points that were computed using measurements corresponding to different beam vectors.  Once $S_1$ is retrieved, the position of the diffuse (two-bounce) spot can also be determined using angle and time-of-flight constraints.  

We encountered this issue when scanning the shape of the large mirror during the experiment reported in Sec. 4.2 of the main text.  The mirror was much taller than it was wide, and was also oriented at an angle such that its surface subtended a relatively small horizontal angle from the perspective of our lidar system.  As a result, only 9 of the 14 beams which directly illuminated the mirror produced three-bounce returns.  We did not compute the additional $S_1$ positions, and so this effectively reduced the number of points that we acquired on the mirror's surface.  One of the reasons that we encountered this phenomenon was that the transmitter-receiver baseline separation of our lidar system was very large ($25.7 cm$) compared to typical laser scanning systems.  The effect would have been less noticeable if a system with a smaller baseline had been used.

\clearpage
%
%
\bibliographystyle{splncs04}
\bibliography{references}